\RequirePackage[T1]{fontenc}
\documentclass[letterpaper, 10 pt, conference]{ieeeconf}
\IEEEoverridecommandlockouts
\usepackage{amsmath,amssymb,amsfonts}
\usepackage{algorithm}
\usepackage{algpseudocode}
\usepackage{booktabs}
\usepackage{multirow}
\usepackage{xcolor}
\usepackage{graphicx}
\usepackage{capt-of}
\usepackage{cite}
\makeatletter
\providecommand{\bstctlcite}[1]{\@bsphack
  \if@filesw\immediate\write\@auxout{\string\citation{#1}}\fi
  \@esphack}
\makeatother
\usepackage[bookmarks=false]{hyperref}
\hypersetup{pdfauthor={Ziqian Wang, Rui Zhang, Jiayu Sun, Yitian Liu, Xingjian Mao, Minqian Wang, Yao Mu},pdftitle={Q-VGM: Q-Guided Value-Gradient Matching for Offline-to-Online RL of Flow-Matching VLA Policies},colorlinks=false,pdfborder={0 0 0}}
\usepackage{cleveref}

\crefname{figure}{Fig.}{Figs.}
\Crefname{figure}{Fig.}{Figs.}
\crefname{table}{Table}{Tables}
\Crefname{table}{Table}{Tables}
\crefname{section}{Section}{Sections}
\Crefname{section}{Section}{Sections}
\crefname{algorithm}{Algorithm}{Algorithms}
\Crefname{algorithm}{Algorithm}{Algorithms}

\renewcommand{\paragraph}[1]{\par\vspace{3pt}\noindent\textbf{#1}\hspace{0.4em}\ignorespaces}

\DeclareMathOperator{\sg}{sg}
\title{\LARGE \bf
Q-VGM: Q-Guided Value-Gradient Matching for\\
Offline-to-Online RL of Flow-Matching VLA Policies
}

\author{Ziqian Wang$^{2}$, Rui Zhang$^{2}$, Jiayu Sun$^{1}$, Yitian Liu$^{1}$, Xingjian Mao$^{1}$, Minqian Wang$^{2}$, Yao Mu$^{1,\dagger}$\\[2pt]
{\normalsize $^{1}$Shanghai Jiao Tong University \quad $^{2}$University of Michigan, Ann Arbor \quad $^{\dagger}$Corresponding author}}
\begin{document}
\bstctlcite{qvgmBibControl}

\IEEEaftertitletext{\vspace{-0.6\baselineskip}\begin{center}
\includegraphics[width=0.96\linewidth]{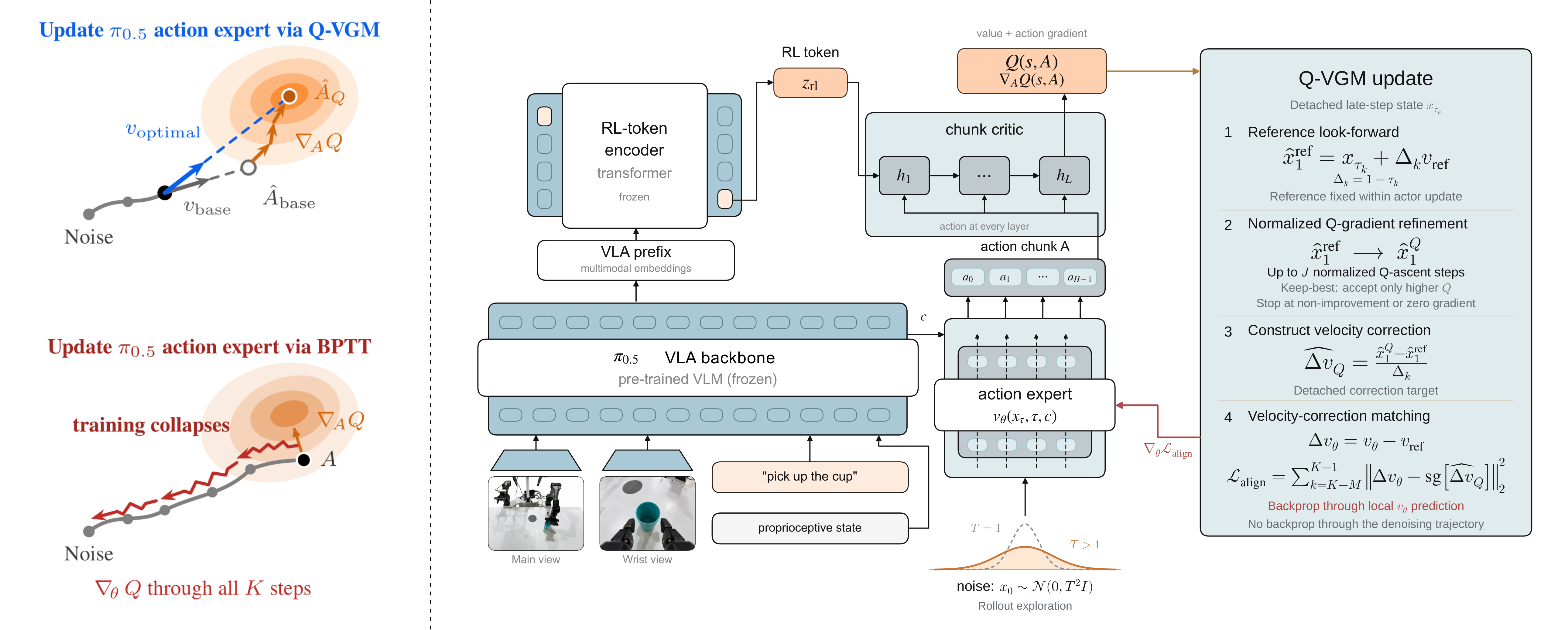}
\captionof{figure}{\textbf{Q-VGM at a glance.} \textit{Left:} Q-VGM moves the action expert along the critic gradient $\nabla_A Q$ through a local velocity target at a late denoising step, whereas backpropagating $\nabla_\theta Q$ through all $K$ denoising steps (BPTT) collapses training. \textit{Right:} a frozen $\pi_{0.5}$ backbone provides an RL token $z_{\mathrm{rl}}$ to an action-sensitive chunk critic $Q(s,A)$; the reference look-forward $\hat x_1^{\mathrm{ref}}$ is refined by up to $J$ normalized $Q$-gradient ascent steps, converted to a velocity target $v_{\mathrm{ref}}+\widehat{\Delta v}_Q$, and matched by the action expert $v_\theta$ with gradients flowing only through $v_\theta$.}
\label{fig:teaser}
\end{center}}
\maketitle

\begin{abstract}
We propose \emph{Q-Guided Value-Gradient Matching} (Q-VGM), an
offline-to-online RL method for fine-tuning flow-matching
vision-language-action policies with a learned critic. Directly applying
critic gradients to flow policies requires backpropagation through the
multi-step denoising process (BPTT), which is costly and unstable at VLA
scale. Motivated by an optimal-control view of denoising, we derive a local approximation connecting clean-action value gradients to local velocity corrections at late denoising steps. Q-VGM uses this connection to construct critic-guided targets for fine-tuning the pretrained action expert through velocity matching. Gradients pass only through local velocity predictions, avoiding backpropagation through the denoising trajectory. An action-sensitive chunk critic is trained with IQL offline and TD
learning online. On LIBERO, Q-VGM improves the few-shot-SFT four-suite average from
84.6\% to 90.7\% offline using 150 policy rollouts per suite, and reaches
98.4\% after online training.
Compared with on-policy RL, Q-VGM achieves \(5.3\times\)
higher sample efficiency on average across four LIBERO suites, measured
by the episode budget for 95\% rollout success rate.
On three real-world bimanual tasks, offline Q-VGM
improves average success from 66.7\% to 98.3\%.
Project page: \url{https://wzzzzq.github.io/qvgm-site/}.
\end{abstract}

\section{Introduction}
\label{sec:introduction}

Vision-language-action (VLA) models bridge high-level multimodal reasoning and
low-level continuous control~\cite{brohan2023rt2,team2024octo,kim2024openvla,black2024pi0,pertsch2025pi05}.
Flow-matching action experts have become their dominant backbone: by
transporting noise to action chunks through iterative denoising, they capture
the multimodal action distributions required for dexterous, long-horizon
manipulation~\cite{lipman2023flow,liu2023flow,black2024pi0,pertsch2025pi05}.

The prevailing post-training recipe for VLA models is supervised fine-tuning
(SFT) on expert demonstrations, which inherits the fundamental limitation of
imitation learning: performance is bounded by the quality and coverage of the
demonstrations, while the abundant failed and suboptimal rollouts a policy
produces carry no learning signal. Reinforcement learning (RL) removes this
bound by directly optimizing task success, allowing a policy to learn from its
own experience and exceed its demonstrator.

Realizing this potential for flow-matching VLAs is difficult. On-policy
methods such as PPO and GRPO require fresh rollouts from the current policy
for each optimization round and discard the collected experience after that
round~\cite{schulman2017ppo,wu2025pirl,ren2025dppo,flowgrpo2025}. Their sample efficiency
is therefore inherently limited, which is tolerable in simulation but becomes
the dominant cost on real robots, where every rollout episode is expensive.
Off-policy value-based RL avoids this inefficiency: all experience is stored
in a replay buffer and repeatedly reused to train a \(Q\)-function~\cite{haarnoja2018sac}. The same
machinery applies when the buffer is fixed in advance, so off-policy training
naturally spans both an offline phase on previously collected data and an
online phase on freshly gathered experience~\cite{nakamoto2023calql}.
To improve the policy, classical
off-policy actor--critic methods such as TD3 and SAC adjust the policy
parameters to ascend the critic along its action gradient
\(\nabla_A Q(s,A)\)~\cite{fujimoto2018td3,haarnoja2018sac}.
For a flow-matching policy, however, this update requires
backpropagating through the entire denoising chain
(backpropagation through time, BPTT)~\cite{wang2023diffusion}.
The resulting gradient contains products of denoising-step
Jacobians, which can amplify or attenuate the critic signal
and destabilize policy optimization~\cite{psenka2024qsm}.
In our VLA fine-tuning experiments, we observe exploding actor
gradients and training collapse under this update
(\Cref{fig:bptt}).

BPTT-free policy extraction has also been explored for generative
policies~\cite{psenka2024qsm,qam2026}.
In our setting, the critic evaluates clean action chunks, whereas
the pretrained VLA generates actions through iterative denoising.
We therefore focus on converting clean-action value gradients
into local velocity supervision, without differentiating through
the denoising trajectory.

To address this challenge, we propose \emph{Q-Guided Value-Gradient
Matching} (Q-VGM), an off-policy fine-tuning method for flow-matching
VLA policies. Motivated by an optimal-control view of denoising, we
start from the relation between optimal velocity corrections and
denoising-time value gradients. We derive a late-step approximation
that estimates these value gradients using clean-action $Q$-gradients
at reference-policy look-forward endpoints. This connection motivates
a velocity-matching objective for fine-tuning the pretrained action
expert.

In practice, Q-VGM refines these endpoint estimates through normalized
$Q$-gradient ascent with keep-best selection. The accepted endpoint
displacements are converted into local velocity targets.
With denoising states and targets detached, backpropagation is
restricted to local velocity predictions rather than the denoising
trajectory. An action-sensitive chunk critic is trained on compact
latent states from the frozen VLA prefix, using IQL offline and
standard TD learning online. At inference time, the policy samples
with the fine-tuned velocity field alone, fully amortizing critic
guidance into the action expert.

On LIBERO,
offline Q-VGM lifts a few-shot-SFT \(\pi_{0.5}\) policy from 84.6\% to 90.7\%
average success across four suites with 150 rollout episodes per suite, and
offline-to-online training reaches 98.4\%, surpassing PPO fine-tuning
(97.4\%). Compared with on-policy RL, Q-VGM achieves \(5.3\times\)
higher sample efficiency on average across four LIBERO suites, measured
by the episode budget for 95\% rollout success rate with exploration noise. On three real bimanual manipulation tasks, offline Q-VGM
raises the average success rate from 66.7\% to 98.3\%.

Our contributions are as follows.

\textbf{(1) A BPTT-free velocity-matching method for VLA fine-tuning.}
Motivated by an optimal-control view of denoising, we derive a
late-step approximation relating clean-action $Q$-gradients to
velocity corrections. This connection motivates a local
velocity-matching objective that directly fine-tunes the pretrained
action expert without backpropagating through the denoising trajectory.

\textbf{(2) An offline-to-online framework for flow-matching VLAs.}
We integrate critic-guided velocity matching with an action-sensitive
chunk critic trained from a replay buffer. The framework uses the same actor
objective across offline and online learning, with phase-specific
critic backups and reference-policy updates.

\textbf{(3) Empirical validation in simulation and on real robots.}
On LIBERO, we demonstrate offline policy improvement and more
sample-efficient online fine-tuning than PPO. We further validate
offline Q-VGM on three real-world bimanual manipulation tasks.

\section{Related Work}
\label{sec:related_work}

\paragraph{RL for VLA policies.}
PPO- and GRPO-style updates have been adapted to diffusion and
flow policies through stochastic formulations or surrogate
objectives~\cite{wu2025pirl,ren2025dppo,reinflow2025,
fpo2025,flowgrpo2025}.
A complementary line uses advantage-based supervised policy
improvement~\cite{peng2019awr}.
For VLAs, RECAP~\cite{pistar2025recap} incorporates
advantage-derived optimality indicators into supervised
policy training.
Alternatively, RL can optimize lightweight policies built
on frozen VLA models.
RLT~\cite{xu2026rlt} conditions a lightweight actor on compact
VLA representations and reference actions, while
PLD~\cite{xiao2025pld} learns residual policies and subsequently
distills collected trajectories back into the VLA through SFT.
In both cases, the RL updates operate on auxiliary policies
rather than the pretrained action expert.

\paragraph{Value-guided optimization of diffusion and flow policies.}
Inference-time methods rank action candidates using value
estimates, learned verifiers, or confidence
scores~\cite{nakamoto2024vgps,song2025hume,
kwok2025robomonkey,mgselect2026}.
QFQL~\cite{jang2025qfql} guides flow sampling at inference
by adding critic gradients evaluated at intermediate noisy
actions to the velocity field.
These methods leave policy parameters unchanged.

At training time, Diffusion-QL~\cite{wang2023diffusion}
combines denoising behavior cloning with $Q$-maximization
through backpropagation along the full denoising chain.
EDP~\cite{kang2023edp} reduces this cost through approximate
action reconstruction, while DTQL~\cite{chen2024dtql}
and FQL~\cite{park2025fql} optimize separate one-step actors
with diffusion- and flow-based behavior regularization,
respectively.

Q-score matching~\cite{psenka2024qsm} instead provides
denoising-time supervision without full-chain differentiation.
Optimal-control and value-gradient formulations offer
related perspectives on reward-guided
fine-tuning~\cite{uehara2024finetuning,tang2024soc,
domingo2025adjoint,liu2025vggflow}.
QAM~\cite{qam2026} uses adjoint matching for critic-guided
policy extraction, while TRQAM~\cite{trqam2026} adds a
path-space KL trust region.
They study compact flow policies on state-based control
benchmarks. Adjoint target construction still requires
step-wise vector--Jacobian products through the reference
velocity field, which can be computationally costly when
scaling to large pretrained VLA action experts.

\paragraph{Offline-to-online reinforcement learning.}
Offline-to-online RL uses previously collected experience
to initialize policies and value functions before further
improvement through online
interaction~\cite{kostrikov2022iql,nakamoto2023calql}.
Flow-based methods such as FQL and QAM also study this
setting on continuous-control benchmarks using compact
policy networks rather than pretrained
VLAs~\cite{park2025fql,qam2026}.

\section{Preliminaries}
\label{sec:preliminaries}

\subsection{Flow Matching for VLA Models}
\label{sec:flow_vla}

We use \(\pi_{0.5}\), a flow-based vision-language-action model
~\cite{black2024pi0,pertsch2025pi05}. Given a VLA
conditioning context \(c\), the action expert generates an action chunk
\(A=[a_0,a_1,\ldots,a_{H-1}]\). We set \(\tau=0\) for Gaussian noise and
\(\tau=1\) for the clean action chunk. Given noise
\(\epsilon\sim\mathcal N(0,I)\), the linear flow path is
\[
    x_\tau=(1-\tau)\epsilon+\tau A,\qquad \tau\in[0,1].
\]
The action expert predicts a velocity field \(v_\theta(x_\tau,\tau,c)\)
trained by conditional flow matching~\cite{lipman2023flow}.
At inference time, sampling integrates
this velocity over \(0=\tau_0<\tau_1<\cdots<\tau_K=1\), starting from
\(x_0\sim\mathcal N(0,I)\).

We hold the conditioning context fixed and omit it below.

\subsection{Value-Gradient Guidance for Flow Models}
\label{sec:control}

Let \(v_{\mathrm{ref}}\) denote the velocity field of a reference policy.
We consider policy improvement through a velocity correction
\(\Delta v\) added to the reference denoising dynamics:
\[
    \dot x_\tau=v_{\mathrm{ref}}(x_\tau,\tau)+\Delta v(x_\tau,\tau).
\]
Let \(x_1^{\mathrm{ref}}\) denote the endpoint reached from \(x_\tau\)
at time \(1\) when \(\Delta v=0\). Given a terminal score \(r(x_1)\), we
seek a correction that improves this score while penalizing deviations
from \(v_{\mathrm{ref}}\).
For \(x_\tau=x\), define the denoising-time value function
\begin{equation}\label{eq:value}
 V(x,\tau)=\sup_{\Delta v}\left[r(x_1)-\frac{\rho}{2}
 \int_\tau^1\|\Delta v(x_u,u)\|_2^2\,du\right],
\end{equation}
where the trajectory follows the controlled dynamics and \(\rho>0\)
controls the penalty strength. Thus, \(V(x,\tau)\) is the best regularized
terminal score reachable from the intermediate state \(x\).

When \(V\) is continuously differentiable, it satisfies the
Hamilton--Jacobi--Bellman (HJB) equation~\cite{liberzon2012calculus}:
\begin{equation}\label{eq:hjb}
\begin{aligned}
0={}&\partial_\tau V+\nabla_xV^\top v_{\mathrm{ref}}\\
 &+\max_{\Delta v}\left\{\nabla_xV^\top\Delta v-\frac{\rho}{2}\|\Delta v\|_2^2\right\}.
\end{aligned}
\end{equation}
Maximizing the quadratic term with respect to \(\Delta v\) gives the optimal
correction~\cite{liu2025vggflow}:
\begin{equation}\label{eq:optimal_delta_v}
    \Delta v^\star(x,\tau)=\frac{1}{\rho}\nabla_xV(x,\tau).
\end{equation}
At the clean endpoint, the remaining control cost vanishes, giving
\(V(x_1,1)=r(x_1)\) and
\(\nabla_xV(x_1,1)=\nabla_{x_1}r(x_1)\).

\begin{algorithm*}[t]
\footnotesize
\caption{Q-VGM: offline (top) and offline-to-online (bottom). The critic and
buffer persist across the transition; only the backup target switches from
\eqref{eq:offline_td} to \eqref{eq:online_td}.
$Q$ denotes the mean of current critic heads, and $\bar Q$ the minimum of target heads.}
\label{alg:qvgm}
\newcommand{\algcmt}[1]{\hfill\textcolor{gray}{$\triangleright$\ #1}}
\raggedright
\textbf{Offline Q-VGM}\\[3pt]
1:~$\mathcal D\gets$ rollouts of $\pi_{\theta_0}$ using~\eqref{eq:exploration}
\algcmt{exploration: latent temperature $T$, flow noise scale $\sigma$}\\
2:~$\{Q_{\phi_i}\}\gets$ critic trained on $\mathcal D$ with the IQL
backup~\eqref{eq:offline_td};\;
$\theta_{\mathrm{ref}}\gets\theta_0$
\algcmt{$v_{\mathrm{ref}}=v_{\theta_{\mathrm{ref}}}$}\\
3:~\textbf{for} each gradient step \textbf{do}\\
4:~\quad denoise with $\theta$ (no grad): $s\sim\mathcal D$;\;
$x_0\!\sim\mathcal N(0,I)$;\;
$x_{\tau_{k+1}}\gets x_{\tau_k}+(\tau_{k+1}-\tau_k)\,v_\theta(x_{\tau_k},\tau_k,c)$\\
5:~\quad \textbf{for} each late denoising step $k=K-M,\ldots,K-1$ \textbf{do}
\algcmt{$\Delta_k=1-\tau_k$}\\
6:~\quad\quad ${\color{red}\hat x_1^{\mathrm{ref}}\gets x_{\tau_k}+\Delta_k\,v_{\mathrm{ref}}(x_{\tau_k},\tau_k,c)}$
\algcmt{one-step look-forward with the reference policy}\\
7:~\quad\quad $\hat x_1^Q\gets$ improve $\hat x_1^{\mathrm{ref}}$ by normalized $Q$-gradient ascent
\algcmt{keep-best gate; freeze at first non-improvement}\\
8:~\quad\quad $\ell_k\gets$
$\big\|\Delta v_\theta(x_{\tau_k},\tau_k,c)-{\color{red}\sg\big[\tfrac{\hat x_1^Q-\hat x_1^{\mathrm{ref}}}{\Delta_k}\big]}\big\|_2^2$\\
9:~\quad \textbf{end for}\\
10:~\quad update $\theta$ once to minimize $\mathcal L_{\mathrm{align}}=\sum_{k=K-M}^{K-1}\ell_k$
\algcmt{gradients only through local $v_\theta$}\\
11:~\textbf{end for}
\algcmt{critic discarded at inference}\\[8pt]
\hrule
\vspace{5pt}
\textbf{Offline-to-Online Q-VGM}\\[3pt]
1:~$\theta$, $\{Q_{\phi_i}\}$, $\mathcal D$ from the offline phase;\; $\theta_{\mathrm{ref}}\gets\theta$
\algcmt{backup switches to~\eqref{eq:online_td}}\\
2:~\textbf{for} each cycle \textbf{do}\\
3:~\quad\textit{(i) Rollout:} collect $B$ episodes with exploration noise, push to $\mathcal D$:\\
4:~\quad\quad $A\gets x_1$ from the noisy flow rollout in~\eqref{eq:exploration}
\algcmt{latent temperature $T$ and per-step flow noise}\\
5:~\quad\textit{(ii) Critic:} $E$ TD epochs on minibatches $(s,A,r,s',d)\sim\mathcal D$:\\
6:~\quad\quad $A'\gets\mathrm{ODE}_\theta(s')$
\algcmt{re-sampled from the current policy (no grad)}\\
7:~\quad\quad $y\gets r+\gamma^{H}(1-d)\bar Q(s',A')$\\
8:~\quad\quad update $\phi$ to minimize $\sum_i\big(Q_{\phi_i}(s,A)-y\big)^2$
\algcmt{targets $\bar\phi$ follow $\phi$ by EMA}\\
9:~\quad\textit{(iii) Actor} (BPTT-free).\;
$Q:=\operatorname{mean}_i Q_{\phi_i}$\\
10:~\quad\;\, denoise with $\theta$ (no grad): $s\sim\mathcal D$;\;
$x_0\!\sim\mathcal N(0,I)$;\;
$x_{\tau_{k+1}}\gets x_{\tau_k}+(\tau_{k+1}-\tau_k)\,v_\theta(x_{\tau_k},\tau_k,c)$\\
11:~\quad\;\, \textbf{for} each late denoising step $k=K-M,\ldots,K-1$ \textbf{do}
\algcmt{$\Delta_k=1-\tau_k$}\\
12:~\quad\quad\;\, ${\color{red}\hat x_1^{\mathrm{ref}}\gets x_{\tau_k}+\Delta_k\,v_{\mathrm{ref}}(x_{\tau_k},\tau_k,c)}$
\algcmt{one-step look-forward with the reference policy}\\
13:~\quad\quad\;\, $\hat x_1^Q\gets$ improve $\hat x_1^{\mathrm{ref}}$ by normalized $Q$-gradient ascent
\algcmt{keep-best gate; freeze at first non-improvement}\\
14:~\quad\quad\;\, $\ell_k\gets$
$\big\|\Delta v_\theta(x_{\tau_k},\tau_k,c)-{\color{red}\sg\big[\tfrac{\hat x_1^Q-\hat x_1^{\mathrm{ref}}}{\Delta_k}\big]}\big\|_2^2$\\
15:~\quad\;\, \textbf{end for}\\
16:~\quad\;\, update $\theta$ once to minimize $\mathcal L_{\mathrm{align}}=\sum_{k=K-M}^{K-1}\ell_k$
\algcmt{gradients only through local $v_\theta$}\\
17:~\quad\textit{(iv) Reference update:} $\theta_{\mathrm{ref}}\gets$ EMA of $\theta$
\algcmt{slowly tracks the current policy}\\
18:~\textbf{end for}
\end{algorithm*}

\section{Method}
\label{sec:method}

We fine-tune a pretrained flow-matching VLA using a learned critic \(Q\),
a replay buffer \(\mathcal D\), and a reference velocity field
\(v_{\mathrm{ref}}\) held fixed within each actor update.
We first describe critic-guided velocity matching
(\Cref{sec:vgm}), followed by the action-sensitive critic
(\Cref{sec:critic}) and the offline-to-online framework
(\Cref{sec:framework}).

\subsection{Critic-Guided Velocity Matching}
\label{sec:vgm}

We instantiate the control objective with terminal score
\(r(x_1)=Q(s,x_1)\), holding the robot state \(s\) fixed throughout
denoising. The actor-induced velocity correction relative to the reference
is \(\Delta v_\theta=v_\theta-v_{\mathrm{ref}}\).
Actor updates directly optimize the pretrained action expert \(v_\theta\).
The terminal condition becomes
\begin{equation}\label{eq:boundary}
    V(x_1,1)=Q(s,x_1),\quad
    \nabla_xV(x_1,1)=\nabla_AQ(s,x_1).
\end{equation}
The optimal correction requires the value gradient at intermediate
denoising states, whereas the critic is trained to evaluate clean action
chunks. We connect these quantities through \(x_1^{\mathrm{ref}}\).

\paragraph{Estimating the value gradient with the critic.}
With \(\dot x_u=v_{\mathrm{ref}}(x_u,u)\), substituting
\eqref{eq:optimal_delta_v} into the HJB equation and applying the chain rule
gives
\begin{equation}\label{eq:reference_derivative}
\begin{aligned}
\frac{d}{du}V(x_u,u)
 &=\partial_uV+\nabla_xV^\top v_{\mathrm{ref}}\\
 &=-\frac{1}{2\rho}\|\nabla_xV(x_u,u)\|_2^2.
\end{aligned}
\end{equation}
Integrating \eqref{eq:reference_derivative} from \(\tau\) to \(1\)
and applying the terminal condition \eqref{eq:boundary} yields
\begin{equation}\label{eq:reference_value}
\begin{aligned}
V(x_\tau,\tau)
={}&Q(s,x_1^{\mathrm{ref}})\\
 &+\frac{1}{2\rho}\int_\tau^1\|\nabla_xV(x_u,u)\|_2^2\,du.
\end{aligned}
\end{equation}
Thus, \(Q(s,x_1^{\mathrm{ref}})\) differs from \(V(x_\tau,\tau)\)
by a remaining-horizon correction. For fixed \(\rho\)
and a locally bounded value gradient, this correction is \(O(1-\tau)\).

For locally smooth \(v_{\mathrm{ref}}\), \(x_1^{\mathrm{ref}}\) admits
the short-horizon expansion
\begin{equation}\label{eq:lookforward}
\begin{aligned}
x_1^{\mathrm{ref}}
 &=x_\tau+\int_\tau^1v_{\mathrm{ref}}(x_u,u)\,du\\
 &=
 \underbrace{x_\tau+(1-\tau)v_{\mathrm{ref}}(x_\tau,\tau)}
 _{\hat x_1^{\mathrm{ref}}}
 +O((1-\tau)^2).
\end{aligned}
\end{equation}

The leading-order term defines the one-step Euler look-forward
estimate \(\hat x_1^{\mathrm{ref}}\).

Under local smoothness with bounded derivatives near \(\tau=1\),
differentiating \eqref{eq:reference_value} and using the short-horizon
expansion in \eqref{eq:lookforward} gives
\begin{equation}\label{eq:gradient_estimate}
\begin{aligned}
\Delta v^\star(x_\tau,\tau)
 &=\frac{1}{\rho}\nabla_xV(x_\tau,\tau)\\
 &=\frac{1}{\rho}\nabla_AQ(s,\hat x_1^{\mathrm{ref}})+O(1-\tau).
\end{aligned}
\end{equation}

The \(O(1-\tau)\) approximation error vanishes as \(\tau\to1\).
We therefore use clean-action \(Q\)-gradients to construct
critic-guided velocity corrections only at the last \(M\) denoising steps,
without differentiating through \(v_{\mathrm{ref}}\).

\paragraph{Refining the Q-gradient correction.}
Let \(\Delta_k=1-\tau_k>0\) denote the remaining denoising time.
Under the local Euler model in \eqref{eq:lookforward},
the leading correction in \eqref{eq:gradient_estimate}
corresponds to a single endpoint update,
\[
\hat x_1^{\mathrm{direct}}
=\hat x_1^{\mathrm{ref}}
+\frac{\Delta_k}{\rho}\nabla_AQ(s,\hat x_1^{\mathrm{ref}}),
\]
whose displacement divided by \(\Delta_k\) recovers the direct velocity
correction.

For a critic learned from limited robot experience, however, the
Q-gradient norm need not indicate how far its local guidance can be
trusted. A fixed gradient coefficient can produce excessive endpoint
displacements into action regions poorly covered by the replay buffer,
making the update sensitive to gradient scale. We therefore refine the
endpoint with small normalized steps, decoupling the update length from
the raw gradient magnitude.

Specifically, we sample \(s\sim\mathcal D\) and
\(x_0\sim\mathcal N(0,I)\), and denoise with the current actor.
At each selected late step, we initialize
\(\hat x_1^{(0)}=\hat x_1^{\mathrm{ref}}\) using
\eqref{eq:lookforward} and propose up to \(J\) normalized
Q-gradient steps:
\[
\hat x_1^{(i+1)}
=\hat x_1^{(i)}
+\alpha\frac{\nabla_AQ(s,\hat x_1^{(i)})}
{\left\|\nabla_AQ(s,\hat x_1^{(i)})\right\|_2},
\]
where \(\alpha>0\) specifies the action-space step length. A keep-best
gate accepts a proposal only if it strictly increases the estimated
\(Q\)-value. Refinement stops at the first non-improving proposal or when
the gradient vanishes. We denote the last accepted endpoint by
\(\hat x_1^Q\), retaining \(\hat x_1^{\mathrm{ref}}\) if no proposal is
accepted. The finite step budget bounds the total endpoint displacement
by \(J\alpha\).

Under the same Euler approximation, a local velocity correction shifts
the predicted endpoint by \(\Delta_k\widehat{\Delta v}_Q\).
Equating this shift to the accepted endpoint displacement gives
\begin{equation}\label{eq:target}
\widehat{\Delta v}_Q(x_{\tau_k},\tau_k)
=\frac{\hat x_1^Q-\hat x_1^{\mathrm{ref}}}{\Delta_k}.
\end{equation}
We train the action expert to match these corrections:
\begin{equation}\label{eq:matching}
\mathcal L_{\mathrm{align}}
=\sum_{k=K-M}^{K-1}
\left\|\Delta v_\theta(x_{\tau_k},\tau_k)
-\widehat{\Delta v}_Q(x_{\tau_k},\tau_k)\right\|_2^2.
\end{equation}
The denoising states and constructed targets are detached, so gradients
flow only through the local actor prediction \(v_\theta\).

\subsection{Action-Sensitive Chunk Critic}
\label{sec:critic}

Critic-guided velocity matching requires reliable action-space gradients
\(\nabla_A Q(s,A)\), where \(A=[a_0,\ldots,a_{H-1}]\) is a continuous
normalized clean-action chunk. Following
Q-chunking~\cite{li2025qchunking}, we use a chunk-level critic trained on
chunk-aligned transitions \((s,A,r,s',d)\) with the discounted in-chunk
return \(r=\sum_{i=0}^{H-1}\gamma^i r_{t+i}\). The phase-specific critic
backups are described in \Cref{sec:framework}.

\paragraph{State representation.}
Following the \(\pi_{0.5}\) architecture, we call the token sequence that the
frozen VLM backbone produces from the visual observation, language
instruction, and proprioceptive state the VLA \emph{prefix}. Feeding all
prefix tokens into the critic is computationally impractical, so we compress
the frozen prefix into an RL token \(z_{\mathrm{rl}}\), adopting the
autoencoder architecture of Xu \emph{et al.}~\cite{xu2026rlt}: the autoencoder
is pretrained by reconstructing the prefix and then frozen, and the critic
consumes \(z_{\mathrm{rl}}\) from the frozen encoder. Because
\(z_{\mathrm{rl}}\) is high-dimensional relative to the action chunk, we
re-inject \(A\) at every hidden layer of the critic, keeping the
\(Q\)-network's action pathway active for reliable \(\nabla_AQ\).

\subsection{Offline-to-Online RL Framework}
\label{sec:framework}

Training proceeds in offline and online stages using a replay buffer
\(\mathcal D\). Both stages use the actor objective
in \eqref{eq:matching}, with phase-specific critic backups and reference
updates (Algorithm~\ref{alg:qvgm}).

\paragraph{Exploratory data collection.}
To populate \(\mathcal D\) with diverse rollouts, collection uses latent
temperature \(T>1\) and Gaussian \emph{flow noise} with scale \(\sigma\):
\begin{equation}\label{eq:exploration}
\begin{aligned}
x_0&\sim\mathcal N(0,T^2I),\\
x_{\tau_{k+1}}&=x_{\tau_k}+\delta_k v_\theta(x_{\tau_k},\tau_k,c)
                  +\sigma\sqrt{\delta_k}\,\varepsilon_k,
\end{aligned}
\end{equation}
where \(\delta_k=\tau_{k+1}-\tau_k\) and
\(\varepsilon_k\sim\mathcal N(0,I)\) independently.
These exploration settings apply only to data collection.
During training, actor-target construction and TD-target sampling use
\(T=1\) and \(\sigma=0\); the same settings apply at evaluation.

\paragraph{Offline training.}
The offline stage uses a fixed buffer containing only exploratory
rollouts of the pretrained policy. We use the fixed pretrained
policy as the reference. Bootstrapping from actions outside the buffer's
support invites value over-estimation. We therefore use the
in-support backup of implicit Q-learning (IQL)~\cite{kostrikov2022iql}: an
action-free value head \(V_{\mathrm{IQL}}(s)\), fit by expectile regression
toward the detached \(Q\)-estimate, provides the target
\begin{equation}\label{eq:offline_td}
y=r+\gamma^H(1-d)V_{\mathrm{IQL}}(s').
\end{equation}

\paragraph{Online training.}
Online training resumes from the offline actor, critic, and replay buffer,
adding new rollouts to \(\mathcal D\) and updating the critic followed by
the actor. We switch to a policy-sampled TD backup to evaluate actions
from the evolving actor. The next action \(A'\) is re-sampled from the current policy by
a clean denoising rollout (no gradient), and
\begin{equation}\label{eq:online_td}
y=r+\gamma^H(1-d)\bar Q(s',A'),
\end{equation}
where \(\bar Q\) is the clipped (min) estimate of the target
\(Q\)-heads~\cite{fujimoto2018td3}.
\\
During online training, we initialize the reference policy from
the offline-trained actor and update its parameters through an
exponential moving average (EMA). This allows the reference to
adapt to the evolving actor rather than remain fixed at the
pretrained checkpoint, while smoothing parameter changes across
successive actor updates. The resulting reference velocity is
used for look-forward estimation and critic-guided velocity matching.
The reference parameters remain fixed within each actor update.

\section{Experiments}
\label{sec:experiments}

In this section, we evaluate Q-VGM by asking the following four research questions:

\textbf{RQ1:}
How much can offline Q-VGM improve a few-shot SFT flow-matching VLA policy using only self-generated rollouts?

\textbf{RQ2:}
Does offline Q-VGM outperform filtered behavior supervision, test-time selection, and test-time guidance under the same rollout data and critic?

\textbf{RQ3:}
Can Q-VGM extend from offline to online training, and how does it compare with online RL fine-tuning in final success rate and sample efficiency?

\textbf{RQ4:}
Does Q-VGM work on real bimanual manipulation tasks?

We answer these questions through simulation benchmarks and real-robot deployment.

\subsection{Experimental Setup}
\begin{figure}[t]
\centering
\includegraphics[width=0.95\linewidth]{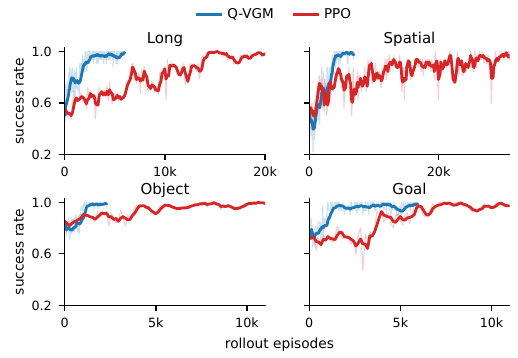}
\caption{Sample efficiency of Q-VGM and PPO on LIBERO. Curves show
one training run per method using the same random seed value and report
the rollout success rate with exploration noise.}
\label{fig:four_task_sample_eff}
\end{figure}

Across all settings, we initialize \(\pi_{0.5}\) with few-shot SFT: on
Spatial, Object and Goal we use the few-shot SFT checkpoint released with \(\pi_{\mathrm{RL}}\)~\cite{wu2025pirl}, and on Long we fine-tune \(\pi_{0.5}\) from the
base model ourselves for 400 steps on 50 demonstrations (5 per task) in
fp32.\footnote{The released checkpoint and training pipeline use bf16
master weights; on the long-horizon suite we found this insufficient for
any method to make progress, so all Long experiments (SFT and every RL
method) are run in fp32.} All RL methods are trained by
us from this initialization under identical conditions.
For the LIBERO comparison, we run all offline and online methods with
three independent training seeds to assess robustness across training runs.
We evaluate each policy without exploration noise using 50 distinct
evaluation seeds per task, with one rollout per seed (500 episodes per
suite). The learning curves in
\Cref{fig:bptt,fig:four_task_sample_eff} instead report the rollout
success rate with exploration noise.

\paragraph{Offline.}
The SFT policy collects 150 rollout episodes per suite with the exploration
noise of \Cref{alg:qvgm}. These rollouts alone form the fixed buffer for
critic training and policy improvement, with no further interaction.
No demonstrations are added to the RL buffer.

\paragraph{Offline-to-online.}
Training resumes from the offline policy, critic, and buffer, and collection
is switched on, following the cycle of \Cref{alg:qvgm}.

\paragraph{Baselines.}
We compare against two groups, all starting from the same SFT checkpoint.
The \emph{offline group} uses the same rollout buffer and, where a critic is
needed, shares ours: \emph{Filtered BC}$^{*}$, flow-matching imitation of the
successful rollouts; \emph{RECAP}, advantage-conditioned
self-training~\cite{pistar2025recap}; \emph{Best-of-\(N\)}, which samples
\(N{=}8\) action chunks at inference and executes the highest-valued one
under the critic~\cite{nakamoto2024vgps,song2025hume}; and
\emph{test-time \(Q\) guidance (QFQL)}~\cite{jang2025qfql},
which applies QFQL's inference-time guidance rule to the frozen
SFT policy using our shared offline IQL critic. The \emph{online group} interacts
with the environment: \emph{Diffusion-QL}, which adds a \(Q\)-maximization
term backpropagated through the denoising chain to the diffusion BC
loss~\cite{wang2023diffusion} (BC + BPTT); its \emph{BPTT-only} variant,
which drops the BC term and is thus the pure critic-gradient update of
TD3/SAC applied through the sampler (gradients reach the action expert,
matching the trainable scope of Q-VGM); \emph{Filtered BC}$^{*}$,
online self-imitation on successful rollouts; and
\emph{PPO (flow-SDE)}, on-policy RL with a stochastic-flow likelihood
surrogate~\cite{wu2025pirl}.

\paragraph{Implementation details.}
For online LIBERO training, we use sparse rewards and an EMA reference
policy across Spatial, Object, Goal, and Long. \Cref{tab:hyperparameters}
summarizes key settings, including offline IQL and autoencoder pretraining.
EMA coefficients denote the weight assigned to the current parameters.

\begin{table}[!ht]
\centering
\caption{Key hyperparameters for Q-VGM in simulation and on real robots.}
\label{tab:hyperparameters}
\footnotesize
\setlength{\tabcolsep}{2pt}
\renewcommand{\arraystretch}{1.0}
\begin{tabular*}{\linewidth}{@{\extracolsep{\fill}}lcc@{}}
\toprule
Setting & LIBERO & Real robot \\
\midrule
Chunk length $H$ & 5 & 10 \\
Action dimension & 7 & 14 \\
Q-head count & 10 & 10 \\
Q-head widths & [512, 512, 256] & [512, 512, 256] \\
RGB inputs & 2 views & 3 views \\
Camera views & Wrist, third view & Front, left/right wrist \\
Control & Delta EEF & Relative joint pose \\
\midrule
\end{tabular*}
\par\nointerlineskip
\begin{tabular*}{\linewidth}{@{\extracolsep{\fill}}lr@{\hspace{12pt}}lr@{}}
\multicolumn{2}{@{}l}{\textbf{Shared Q-VGM}} &
\multicolumn{2}{l@{}}{\textbf{Offline IQL}} \\
\midrule
Denoising steps $K$ & 10 & Training steps & 6,000 \\
Guided late steps $M$ & 5 & Batch (frozen enc.) & 64 \\
Q-ascent steps $J$ & 3 & Expectile $\tau_{\mathrm{IQL}}$ & 0.8 \\
Ascent step $\alpha$ & 0.05 & Q LR & $1\times10^{-4}$ \\
Actor grad. clip & 3.5 & Value LR & $1\times10^{-4}$ \\
Discount $\gamma$ & 0.99 & Critic target EMA & 0.005 \\
\midrule
\multicolumn{2}{@{}l}{\textbf{Online LIBERO}} &
\multicolumn{2}{l@{}}{\textbf{RL-token autoencoder}} \\
\midrule
Actor LR & $5\times10^{-6}$ & Training steps & 10,000 \\
Critic LR & $3\times10^{-4}$ & Batch size & 32 \\
Global batch & 512 & Learning rate & $1\times10^{-4}$ \\
Actor micro-batch & 8 & Encoder layers & 2 \\
Critic updates/actor & 25 & Decoder layers & 2 \\
Action pert. std. & 0.05 & Token dimension & 2048 \\
Reference EMA & 0.002 & Attention heads & 8 \\
\midrule
\end{tabular*}
\par\nointerlineskip
\begin{tabular*}{\linewidth}{@{\extracolsep{\fill}}lcc@{}}
Online suite setting & Spatial/Object/Goal & Long \\
\midrule
Latent temperature $T$ & 2.0 & 1.5 \\
Flow noise scale $\sigma$ & 0.08 & 0.15 \\
Critic target EMA & 0.005 & 0.02 \\
\bottomrule
\end{tabular*}
\end{table}

\subsection{Simulation Evaluation}
\label{sec:simulation}

This section addresses RQ1--RQ3 on the LIBERO Spatial, Object, Goal and Long suites~\cite{liu2023libero}.

\begin{table}[t]
\centering
\caption{LIBERO success rate (\%) without exploration noise (50 eps/task,
500 eps/suite). Offline and online methods: mean $\pm$ std over three training seeds.}
\label{tab:simulation_results}
\footnotesize
\setlength{\tabcolsep}{1.5pt}
\begin{tabular*}{\linewidth}{@{}l@{\extracolsep{\fill}}ccccc@{}}
\toprule
Method & Spatial & Object & Goal & Long & Avg \\
\midrule
\multicolumn{6}{@{}l}{\textit{Full-dataset SFT~\cite{wu2025pirl}}} \\
Octo~\cite{team2024octo} & 78.9 & 85.7 & 84.6 & 51.1 & 75.1 \\
OpenVLA~\cite{kim2024openvla} & 84.7 & 88.4 & 79.2 & 53.7 & 76.5 \\
OpenVLA-OFT~\cite{kim2025oft} & 91.6 & 95.3 & 90.6 & 86.5 & 91.0 \\
\(\pi_0\)-FAST~\cite{pertsch2025fast} & 96.4 & 96.8 & 88.6 & 60.2 & 85.5 \\
\(\pi_0\)~\cite{black2024pi0} & 96.8 & 98.8 & 95.8 & 85.2 & 94.2 \\
\(\pi_{0.5}\)~\cite{pertsch2025pi05} & 98.8 & 98.2 & 98.0 & 92.4 & 96.9 \\
\midrule
\multicolumn{6}{@{}l}{\textit{Few-shot SFT (initialization)}} \\
\(\pi_{0.5}\) few-shot SFT & 77.6 & 97.0 & 86.2 & 77.6 & 84.6 \\
\midrule
\multicolumn{6}{@{}l}{\textit{Few-shot SFT + offline RL (150 rollout episodes per suite)}} \\
Filtered BC$^{*}$ &
78.3\kern0.4pt{\fontsize{6}{7}\selectfont\(\pm\)0.7} &
97.1\kern0.4pt{\fontsize{6}{7}\selectfont\(\pm\)0.9} &
86.5\kern0.4pt{\fontsize{6}{7}\selectfont\(\pm\)1.8} &
78.5\kern0.4pt{\fontsize{6}{7}\selectfont\(\pm\)1.1} &
85.1\kern0.4pt{\fontsize{6}{7}\selectfont\(\pm\)0.6} \\
RECAP~\cite{pistar2025recap} &
79.9\kern0.4pt{\fontsize{6}{7}\selectfont\(\pm\)2.1} &
97.5\kern0.4pt{\fontsize{6}{7}\selectfont\(\pm\)0.8} &
87.8\kern0.4pt{\fontsize{6}{7}\selectfont\(\pm\)0.9} &
79.1\kern0.4pt{\fontsize{6}{7}\selectfont\(\pm\)2.4} &
86.1\kern0.4pt{\fontsize{6}{7}\selectfont\(\pm\)0.7} \\
Best-of-\(N\) (\(N{=}8\)) &
80.1\kern0.4pt{\fontsize{6}{7}\selectfont\(\pm\)2.0} &
97.0\kern0.4pt{\fontsize{6}{7}\selectfont\(\pm\)0.9} &
87.5\kern0.4pt{\fontsize{6}{7}\selectfont\(\pm\)1.8} &
81.3\kern0.4pt{\fontsize{6}{7}\selectfont\(\pm\)1.3} &
86.5\kern0.4pt{\fontsize{6}{7}\selectfont\(\pm\)1.1} \\
QFQL (test-time)~\cite{jang2025qfql} &
84.5\kern0.4pt{\fontsize{6}{7}\selectfont\(\pm\)1.9} &
97.1\kern0.4pt{\fontsize{6}{7}\selectfont\(\pm\)0.3} &
88.2\kern0.4pt{\fontsize{6}{7}\selectfont\(\pm\)1.7} &
80.2\kern0.4pt{\fontsize{6}{7}\selectfont\(\pm\)2.3} &
87.5\kern0.4pt{\fontsize{6}{7}\selectfont\(\pm\)1.2} \\
\textbf{Q-VGM} (offline) &
\textbf{88.9}\kern0.4pt{\fontsize{6}{7}\selectfont\(\pm\)1.4} &
\textbf{98.1}\kern0.4pt{\fontsize{6}{7}\selectfont\(\pm\)0.7} &
\textbf{92.2}\kern0.4pt{\fontsize{6}{7}\selectfont\(\pm\)1.2} &
\textbf{83.6}\kern0.4pt{\fontsize{6}{7}\selectfont\(\pm\)1.5} &
\textbf{90.7}\kern0.4pt{\fontsize{6}{7}\selectfont\(\pm\)0.5} \\
\midrule
\multicolumn{6}{@{}l}{\textit{Few-shot SFT + online RL}} \\
Diffusion-QL~\cite{wang2023diffusion} &
71.3\kern0.4pt{\fontsize{6}{7}\selectfont\(\pm\)4.5} &
86.5\kern0.4pt{\fontsize{6}{7}\selectfont\(\pm\)3.4} &
82.7\kern0.4pt{\fontsize{6}{7}\selectfont\(\pm\)3.9} &
70.1\kern0.4pt{\fontsize{6}{7}\selectfont\(\pm\)5.4} &
77.7\kern0.4pt{\fontsize{6}{7}\selectfont\(\pm\)3.2} \\
Diffusion-QL {\fontsize{6}{7}\selectfont (BPTT only)} &
33.4\kern0.4pt{\fontsize{6}{7}\selectfont\(\pm\)8.3} &
41.1\kern0.4pt{\fontsize{6}{7}\selectfont\(\pm\)10.2} &
36.9\kern0.4pt{\fontsize{6}{7}\selectfont\(\pm\)9.3} &
32.8\kern0.4pt{\fontsize{6}{7}\selectfont\(\pm\)10.2} &
36.1\kern0.4pt{\fontsize{6}{7}\selectfont\(\pm\)5.6} \\
Filtered BC$^{*}$ &
84.7\kern0.4pt{\fontsize{6}{7}\selectfont\(\pm\)1.9} &
99.1\kern0.4pt{\fontsize{6}{7}\selectfont\(\pm\)0.6} &
92.4\kern0.4pt{\fontsize{6}{7}\selectfont\(\pm\)1.6} &
85.0\kern0.4pt{\fontsize{6}{7}\selectfont\(\pm\)2.2} &
90.3\kern0.4pt{\fontsize{6}{7}\selectfont\(\pm\)0.8} \\
PPO (flow-SDE)~\cite{wu2025pirl} &
98.9\kern0.4pt{\fontsize{6}{7}\selectfont\(\pm\)0.6} &
99.9\kern0.4pt{\fontsize{6}{7}\selectfont\(\pm\)0.1} &
97.7\kern0.4pt{\fontsize{6}{7}\selectfont\(\pm\)0.9} &
93.1\kern0.4pt{\fontsize{6}{7}\selectfont\(\pm\)0.8} &
97.4\kern0.4pt{\fontsize{6}{7}\selectfont\(\pm\)0.3} \\
\textbf{Q-VGM} (online) &
\textbf{99.8}\kern0.4pt{\fontsize{6}{7}\selectfont\(\pm\)0.2} &
\textbf{99.9}\kern0.4pt{\fontsize{6}{7}\selectfont\(\pm\)0.1} &
\textbf{99.4}\kern0.4pt{\fontsize{6}{7}\selectfont\(\pm\)0.6} &
\textbf{94.6}\kern0.4pt{\fontsize{6}{7}\selectfont\(\pm\)0.3} &
\textbf{98.4}\kern0.4pt{\fontsize{6}{7}\selectfont\(\pm\)0.2} \\
\bottomrule
\end{tabular*}
\par\vspace{2pt}
{\footnotesize\raggedright $^{*}$Filtered BC imitates successful policy-generated trajectories.\par}
\end{table}

\paragraph{Offline results (RQ1, RQ2).}
\Cref{tab:simulation_results} shows that offline Q-VGM raises the four-suite average from 84.6\% to 90.7\% using only 150 noisy rollout episodes per suite
and no additional expert data, outperforming every offline baseline.
Behavior supervision on the same rollouts (Filtered BC 85.1\%, RECAP 86.1\%)
recovers little, as it is bounded by the behaviors the policy already
produces, and Best-of-\(N\) (86.5\%) shows that re-ranking policy samples is
similarly limited.
Test-time \(Q\) guidance (QFQL) achieves 87.5\% by querying
the critic at intermediate noisy states during denoising and
directly using the resulting action gradients to guide sampling.

\begin{figure}[t]
\centering
\includegraphics[width=\linewidth]{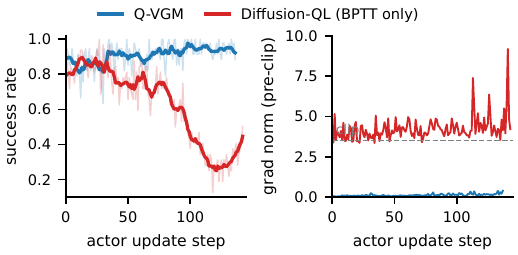}
\caption{Q-VGM vs.\ Diffusion-QL (BPTT only) on LIBERO-Long task 2: rollout
success rate with exploration noise (left) and actor gradient norm before
clipping (right).}
\label{fig:bptt}
\end{figure}

\paragraph{Online results (RQ3).}
Across three training seeds, online Q-VGM achieves a four-suite average
success rate of \(98.4 \pm 0.2\%\).
This exceeds PPO with flow-SDE (97.4\%) and the \(\pi_{0.5}\) policy
fine-tuned on the full demonstration dataset (96.9\%;
\Cref{tab:simulation_results}). Diffusion-QL with its BC term achieves
77.7\%, below the SFT initialization; removing the BC term (BPTT only)
further reduces success to 36.1\%. In \Cref{fig:bptt}, the BPTT-only
update collapses, with actor gradient norms an order of magnitude above
Q-VGM's.

\paragraph{Sample efficiency (RQ3).}
\Cref{tab:sample_efficiency} reports episode budgets for 95\% rollout
success rate with exploration noise on four LIBERO task suites. PPO uses
\(4.43\)--\(6.14\times\) as many episodes;
the arithmetic mean of the four ratios is \(5.35\times\).

\begin{table}[t]
\centering
\caption{Episode budgets for 95\% rollout success rate. Both methods report mean episode budgets over three training seeds. Q-VGM includes the 150 offline rollout episodes per suite.}
\label{tab:sample_efficiency}
\footnotesize
\setlength{\tabcolsep}{7pt}
\begin{tabular}{lrrr}
\toprule
Suite & Q-VGM & PPO & PPO / Q-VGM \\
\midrule
Spatial & 3,968.7 & 21,269.3 & \(5.36\times\) \\
Object & 1,152.7 & 7,072.0 & \(6.14\times\) \\
Goal & 1,291.3 & 5,717.3 & \(4.43\times\) \\
Long & 2,518.0 & 13,770.7 & \(5.47\times\) \\
\bottomrule
\end{tabular}
\end{table}

\subsection{Real-Robot Evaluation}
\label{sec:real_robot}

\begin{figure}[t]
\centering
\includegraphics[width=0.32\linewidth]{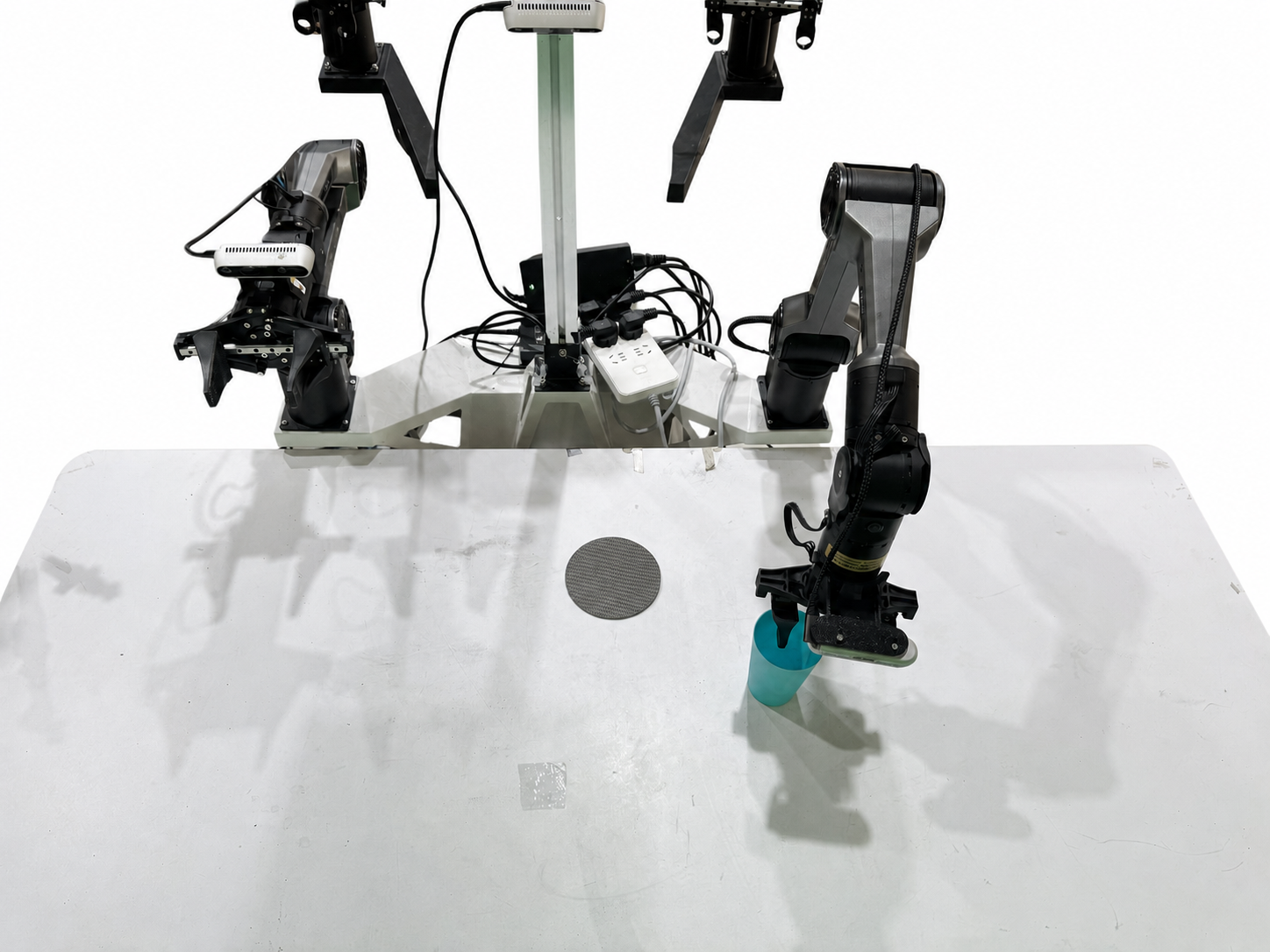}\hfill
\includegraphics[width=0.32\linewidth]{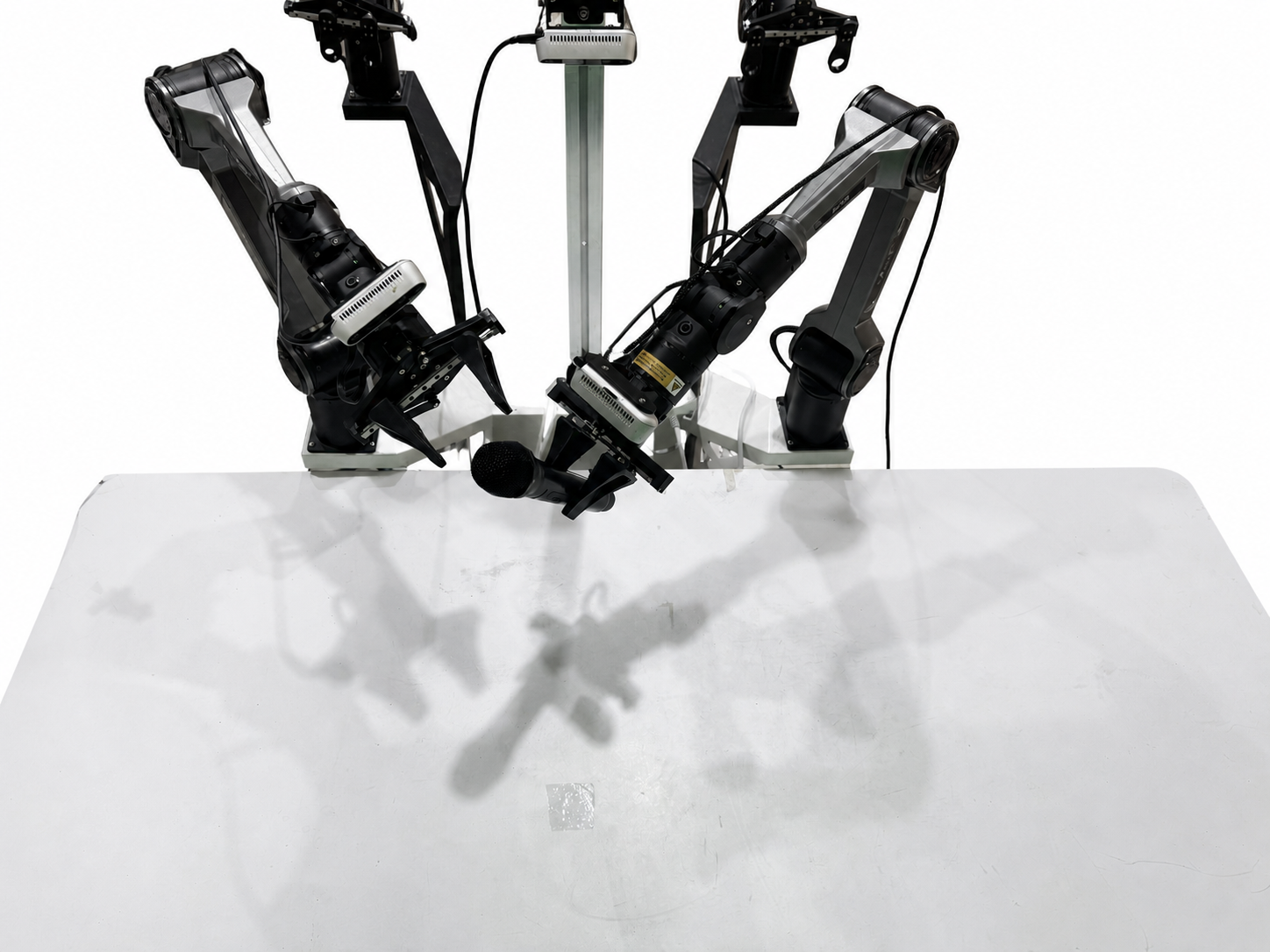}\hfill
\includegraphics[width=0.32\linewidth]{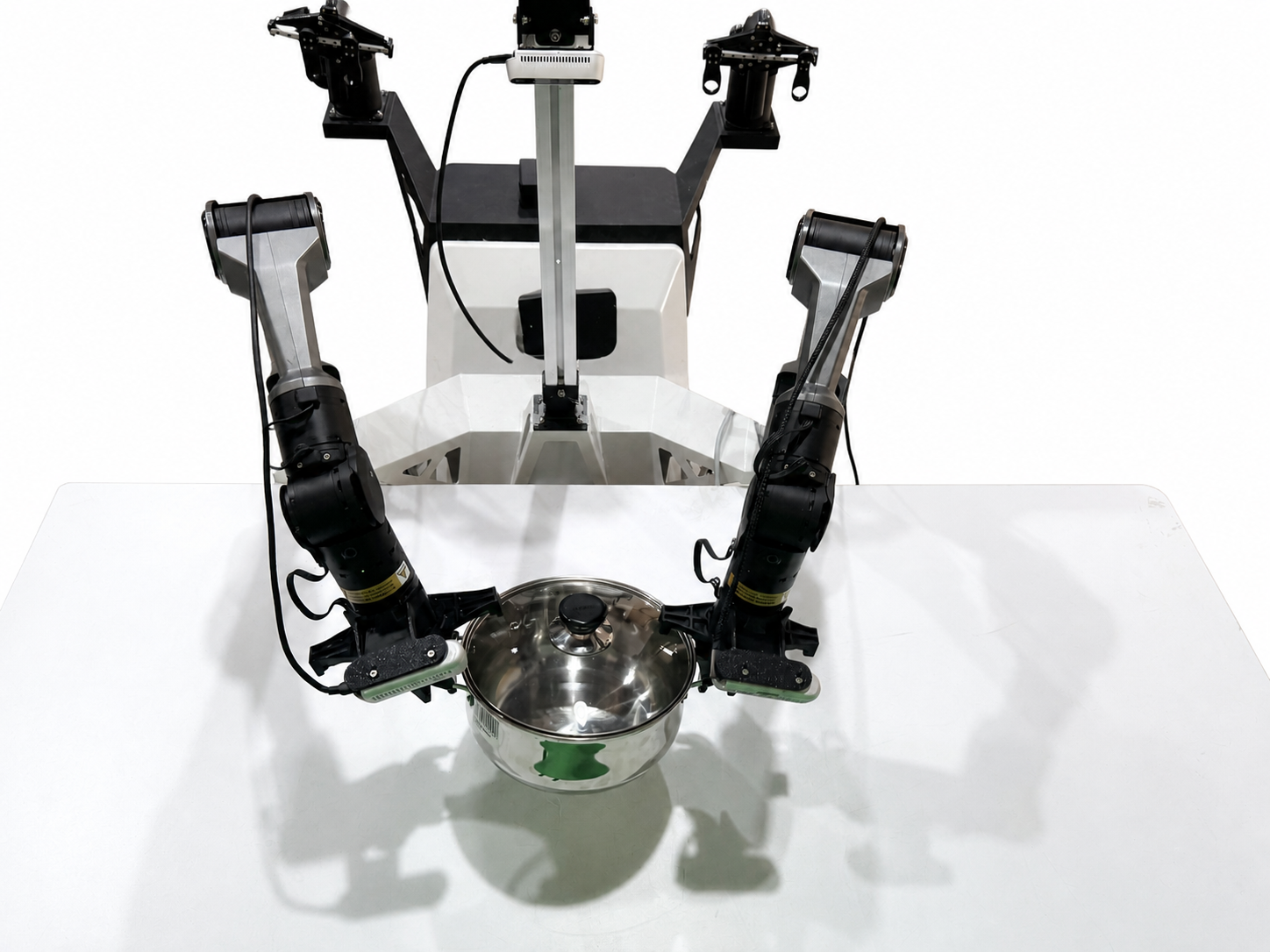}
\caption{Real-robot bimanual manipulation tasks.}
\label{fig:platforms}
\end{figure}

\paragraph{Setup.}
This section addresses RQ4. We evaluate three bimanual manipulation tasks on
a real robot (\Cref{fig:platforms}): Handover Mic, Pick Cup, and Lift Pot.
The \(\pi_{0.5}\) policy is initialized by few-shot SFT with 10
demonstrations per task; the SFT policy then collects 100 rollout episodes
per task for the offline phase, and every method is evaluated over 20 trials
per task on in-distribution initial configurations. The offline baselines
follow the same protocol as in simulation.

\begin{table}[!ht]
\centering
\caption{Real-robot success rates (successes / 20 trials) on Handover Mic,
Pick Cup and Lift Pot. All methods start from the same few-shot SFT policy
(10 demonstrations per task) and use the same 100 rollout episodes per task.}
\label{tab:real_robot}
\footnotesize
\setlength{\tabcolsep}{4pt}
\begin{tabular}{lccc}
\toprule
Method & Mic & Cup & Pot \\
\midrule
SFT & 12/20 & 15/20 & 13/20 \\
Filtered BC$^{*}$ & 15/20 & 17/20 & 15/20 \\
RECAP~\cite{pistar2025recap} & 16/20 & 17/20 & 15/20 \\
Best-of-\(N\) & 15/20 & 16/20 & 18/20 \\
Test-time \(Q\) guidance (QFQL)~\cite{jang2025qfql} & 16/20 & 16/20 & 17/20 \\
\textbf{Q-VGM (offline)} & \textbf{19/20} & \textbf{20/20} & \textbf{20/20} \\
\bottomrule
\end{tabular}
\par\vspace{2pt}
{\footnotesize\raggedright $^{*}$Filtered BC imitates successful policy-generated trajectories.\par}
\end{table}

\paragraph{Results.}
Offline Q-VGM uses the IQL training settings in
\Cref{tab:hyperparameters} and 500 actor updates.
\Cref{tab:real_robot} shows that offline Q-VGM improves success
rates on all three tasks from 12/20, 15/20, and 13/20 to
19/20, 20/20, and 20/20, respectively, outperforming all
offline baselines. These real-robot results, together with
the sample-efficiency gains in simulation, suggest that
Q-VGM's off-policy learning is well suited to real-world
VLA fine-tuning, where robot interactions are costly.

\subsection{Ablation Studies}
\label{sec:ablation}

We ablate the offline critic objective on LIBERO-Spatial, holding the rollout
buffer, RL-token state, and actor update fixed (\Cref{tab:ablation}).

\begin{table}[!ht]
\centering
\caption{Offline critic-objective ablation on LIBERO-Spatial.}
\label{tab:ablation}
\footnotesize
\begin{tabular}{lc}
\toprule
Critic objective & Spatial SR (\%) \\
\midrule
\textbf{IQL (ours)}~\cite{kostrikov2022iql} & \textbf{88.9} \\
CQL~\cite{kumar2020cql} & 87.8 \\
SARSA~\cite{sutton2018rlbook} & 87.0 \\
Cal-QL~\cite{nakamoto2023calql} & 84.5 \\
\bottomrule
\end{tabular}
\end{table}

\Cref{tab:actor_ablation} reports online component ablations on LIBERO-Spatial.
Matching at all denoising steps reduces success from 99.8\% to 96.2\%,
and removing the keep-best gate reduces it to 95.0\%.
Using the current actor for target construction yields 82.0\% and leads
to highly unstable training, with large performance fluctuations.

\begin{table}[!ht]
\centering
\caption{Online Q-VGM component ablations on LIBERO-Spatial.}
\label{tab:actor_ablation}
\footnotesize
\setlength{\tabcolsep}{4pt}
\begin{tabular}{lc}
\toprule
Variant & Spatial SR (\%) \\
\midrule
Full Q-VGM & 99.8 \\
Matching at all denoising steps & 96.2 \\
Without keep-best (fixed $J$ steps) & 95.0 \\
Current actor for target construction & 82.0 \\
\bottomrule
\end{tabular}
\end{table}

\section{Conclusion}
\label{sec:conclusion}

We presented Q-VGM, a sample-efficient off-policy RL method that fine-tunes flow-matching
VLA policies with a learned critic, requiring neither action likelihoods nor
backpropagation through the denoising chain. Casting policy improvement as
optimal control over the denoising dynamics, Q-VGM constructs critic-guided
velocity corrections from look-forward actions to fine-tune the action expert,
and trains a single critic across the offline and online phases with
only the backup target switched. Offline Q-VGM improves a few-shot-SFT \(\pi_{0.5}\) policy from 84.6\% to 90.7\% on LIBERO and from 66.7\% to 98.3\% on three real bimanual manipulation tasks, and offline-to-online training reaches 98.4\%, surpassing PPO fine-tuning. To reach 95\% rollout success across the four LIBERO suites, PPO requires $5.3\times$ as many episodes on average as Q-VGM.

\paragraph{Limitations and future work.}
Q-VGM relies on the critic's action-space gradient, which is most
trustworthy near actions supported by the buffer; keep-best selection falls
back to the base action when ascent does not improve the value, but a
stronger safeguard would internalize a trust region into the sampling
dynamics~\cite{trqam2026}. Scaling the critic to longer horizons and more
diverse environments also remains challenging~\cite{park2025horizon};
combining world models with value estimation, so that a learned dynamics
model shortens the effective TD horizon while the value function keeps
supplying gradient signals, is a promising direction.

\begingroup
\let\qvgmOriginalFootnotesize\footnotesize
\renewcommand{\footnotesize}{\qvgmOriginalFootnotesize\setlength{\baselineskip}{8pt}}
\bibliographystyle{IEEEtran}
\bibliography{references}
\endgroup

\end{document}